# SegET: Deep Neural Network with Rich Contextual Features for Cellular Structures Segmentation in Electron Tomography Image

Enze Zhang[1,2]   Fa Zhang[1]   Zhiyong Liu[1]   Xiaohua Wan[1*]   Lifa Zhu[1,2]

[1]High Performance Computer Research Center, Institute of Computing Technology, Chinese Academy of Sciences, Beijing, China

[2]University of Chinese Academy of Sciences, Beijing, China

{zhangenze, zhangfa, zyliu, wanxiaohua, zhulifa}@ict.ac.cn, * corresponding author

## Abstract

*Electron tomography (ET) allows high-resolution reconstructions of macromolecular complexes at near-native state. Cellular structures segmentation in the reconstruction data from electron tomographic images is often required for analyzing and visualizing biological structures, making it a powerful tool for quantitative descriptions of whole cell structures and understanding biological functions. However, these cellular structures are rather difficult to automatically separate or quantify from view owing to complex molecular environment and the limitations of reconstruction data of ET. In this paper, we propose a single end-to-end deep fully-convolutional semantic segmentation network dubbed SegET with rich contextual features which fully exploits the multi-scale and multi-level contextual information and reduces the loss of details of cellular structures in ET images. We trained and evaluated our network on the electron tomogram of the CTL Immunological Synapse from Cell Image library. Our results demonstrate that SegET can automatically segment accurately and outperform all other baseline methods on each individual structure in our ET dataset.*

## 1. Introduction

Electron Tomography (ET) is a powerful tool to elucidate reconstructions called tomograms of complex biological cellular structures at molecular resolution in close-native state [21]. Precise cellular structures segmentation is critical to accurate analysis of individual complexes in cell tomograms and understanding biological functions [22], as illustrated in Figure 1. In most cases, many cellular structures segmentation tasks from electron tomograms are performed manually by biologists because expert knowledge of the biological constructs as well as the imaging equipment are required. But manual annotation is rather difficult and laborious, as it could take biologists several hours to annotate one single slice of the data stack accurately due to the ever-growing size of the images. Therefore, automatic cellular structures segmentation in ET data is highly desirable. However, this task is even more challenging than other biomedical data segmentation because that individual cellular structures are difficult to identify in the complex molecular environment, in which macromolecules are structurally diverse and densely distributed. Also, reconstruction tomograms suffer from the limitations of ET including the images at a very low signal-to-noise ratio (SNR) and anisotropic resolution caused by missing data [23]. Earlier studies on segmentation approaches for ET images are mainly based on several image processing skills, such as pixel classification by thresholding and the watershed algorithm [1]. But these traditional image processing skills cannot work effectively on the electron tomographic data due to low contrast between target objects and the background [2], as well as differing intensities and orientations. In the last years, there were more approaches using hand-designed features to get segmentation results [3,4,5,6]. Nevertheless, these methods excessively replying on prior knowledge [7,8,9,10] cannot represent high level features well, which will cause the degradation when the dataset is changed or deformed because their capacity to learn the semantic meanings of the whole electron tomograms is low. Recently, several image segmentation methods have been developed with a growing amount of DNNs, especially fully convolutional networks (FCNs) [11]. The features learned from the dataset are more robust and representative since these methods do not require any prior knowledge or hand-crafted features. Badrinarayanan *et al.* [12] proposed SegNet, a deep encoder-decoder network, and it has a good semantic architecture for cellular structures segmentation on fluorescent microscopy Data [13]. Ronneberger *et al.* [14] proposed U-Net, a U-shaped DNN particularly designed for biomedical image segmentation that adds a symmetric expanding path to enable precise localization that the prior methods lack of. This network can be trained end-to-end from very few images. The DeepLabv3+ [15] architecture from DeepLab series is now state-of-the-art for semantic segmentation, extending DeepLabv3 by adding a simple yet effective decoder module to refine the segmentation results.

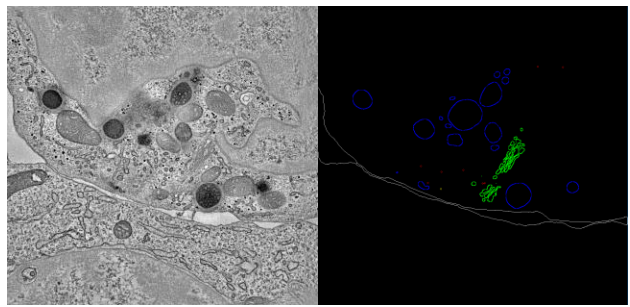

Figure 1. Left: an original slice of an ET image stack. Right: the corresponding segmentation annotation (individual cellular components are denoted by different colors).

However, the SegNet and U-Net both use small-size convolution kernels, causing them struggling in segmenting cellular structures in various sizes, especially large objects. As for FCN and DeepLab, their decoders are way too simple to recover a pretty fine-grained result. They all cannot make fully use of the contextual information of ET images, which contains microcellular details, distinct sizes of cellular objects, blurred structures and borders, anisotropic resolution and too much noise. Therefore, none of these methods can accurately generalize and perform on those cellular structures in ET images. Meanwhile, there is usually very little data to feed a network with a very deep encoder which performs well in classification.

In this work, we present an encoder-center-decoder architecture dubbed SegET with rich contextual features to segment cellular structures in ET images. The main idea is taking fully advantage of the global knowledge from high-level layers to do classification task and local message from low-level layers to figure out the localization problem. Therefore, we focus on improving the classification ability in our encoder and center module and increasing localization accuracy in our decoder module. The network can fully exploit the multi-scale context with different receptive fields and multi-level contextual information from the deep hierarchical structure, thereby reducing the loss of details of cellular structures in ET data and improving the ability of the network to learn the features of objects with various sizes. Mainly three contributions in the architecture are proposed for ET reconstruction images segmentation. First, in the encoder of the network, all the downsampling layers in the encoder part are replaced with the convolution layers with stride 2 to accurately extract features and reduce the loss of detailed information when downsampling. Second, we apply dilated convolution filters with different scales to the encoded features in the center part, and incorporated them together as the final extracted features, thus improving the ability of the network to leverage the multi-scale context with different receptive fields and to learn the features of objects or structures with various sizes. Third, as for the decoder part, we utilize the decoded outputs of all the decoder blocks to fully exploit multi-level contextual information from the deep hierarchical structure. As a result, more precise and fine-grained details of the microcellular structures and borders can be recovered. The proposed method can be trained to segment multiple cellular structures of interest using annotated ground truth data of our ET cell dataset, which performed well compared to our baseline algorithms. We will describe the proposed method in details in the next section.

## 2. Method

### 2.1. Network Architecture of SegET

We present a deep neural network for ET cell segmentation inspired by U-Net. The architecture of the proposed method is illustrated in Figure 2. It basically contains three modules: an encoder module which is a down-sampling path with some convolution layers or maybe max-pooling layers to extract and encode features from input data; a middle module including some convolution layers to integrate and further process the high-level encoded features which have more semantic meanings; and a decoder module defining up-sampling path to recover spatial information from extracted features of input data and to refine the prediction logits.

There are 4 blocks, each of which contains 4 convolution layers with $3\times3$ convolution kernels in the encoder module. The first 2 layers are only designed for feature extraction, the third for feature extraction and channel reduction so as to be concatenated in the skip structure and the last layer with a stride 2 for feature extraction and downsampling. The number of the convolution filters in the first block is 16 and the number of that in every next block doubles in the encoder part.

The center module contains 7 convolution layers with $3\times3$ or $1\times1$ convolution kernels. The first 2 layers performed as high-level feature extractor, followed by 4 parallel conv layers with different dilated sizes for multi-scale receptive field. The last conv layer with $1\times1$ convolution kernels is applied after concatenation of those parallel convolved features to refine the bottleneck features and adjust the channel number.

The decoder still contains 4 basic blocks, including a bilinear upsampling layer with a stride 2 and 2 conv layers with $3\times3$ conv kernels to refine the coarse maps after interpolation. In each block of the decoder module, the coarse map after bilinear upsampling will be concatenated with the skip structure from corresponding encoder block to supplement the details of features lost in encoding process. At the end of the decoder part, we fuse the higher-level decoder features to the output of the last block by concatenating and convolving. The higher-level decoder features also come from the concatenation and convolution of outputs of the higher-level blocks. The final prediction can be regarded as the score map of the last decoder block fused with prior multi-level contextual information activated by a sigmoid function.

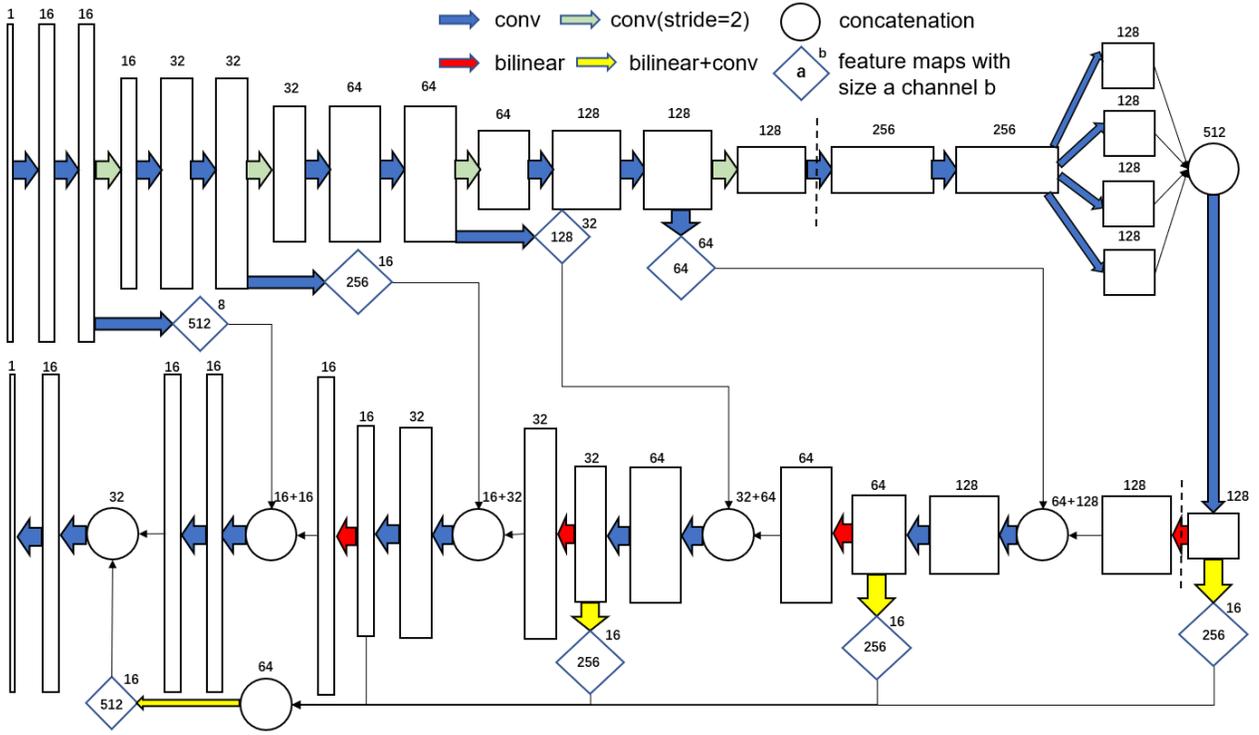

Figure 2. The overview of SegET architecture. The figure above each rectangular, round or diamond shape feature map denotes current number of channels. The module before the first dotted line is the encoder part. The center module includes layers between the first and second dotted line, and the rest parts belong to the decoder.

Besides, Batch Normalization [16] and the rectified linear unit(ReLU) described in [17] are employed after every non-linear operation like convolution, thus helping the convergence of training and improvement of the precision.

### 2.2. Loss function for segmentation

After the activation by the sigmoid function, the output scores are compressed between 0 and 1, representing the possibility of any pixel belonging to foreground object. The training of the whole network can be treated as a per-pixel binary classification with regard to the ground-truth(GT) data labels since our task is to separate one kind of the foreground cellular structures from a series of ET slices. We train the network by minimizing the loss function:

$$L(X:W) = L_b(X:W) + L_j(X:W) + \lambda\psi(W) \quad (1)$$

where $L_b(X:W)$ denotes the binary cross entropy loss with respect to the GT value for pixel x in image, $L_j(X:W)$ denotes the Jaccard distance loss, and the third item of loss function $\lambda\psi(W)$ is a regularizer penalizing the weight parameters from becoming too larger causing overfitting. The tradeoff of the weights between the loss term and the regularizer term is adjusted by coefficient $\lambda$. Additionally, $W$ denotes the parameters of the network for inferring, and $X$ denotes the set of total input image pixel values.

As showed above, we optimize our model by two loss functions together, either representing one kind of optimization idea respectively. The binary cross entropy(BCE) loss is a kind of cross entropy(CE) loss activated by the sigmoid function, regarding to classification with 2 classes. BCE reflects the prediction accuracy of the both foreground and background pixels, expressed as:

$$y - y * t + \log(1 + e^{-y}) \quad (2)$$

Where $t \in GT$, $y \in Y$. $Y$ denotes the set of total output prediction pixel values. To avoid overflow, a stabilized format of BCE loss is expressed as:

$$L_b = \max(y, 0) - y * t + \log(1 + e^{-abs(y)}) \quad (3)$$

Jaccard distance loss derives from Jaccard index, also known as Jaccard similarity coefficient, which is a statistic used for comparing the similarity and diversity of sample sets. The Jaccard coefficient measures similarity between finite sample sets, and is defined as the size of the intersection divided by the size of the union of the sample sets, which expresses as below:

$$J(Y,T) = \frac{|Y \cap T|}{|Y| + |T| - |Y \cap T|} \quad (4)$$

The Jaccard distance, which measures dissimilarity between sample sets, is complementary to the Jaccard coefficient and is obtained by subtracting the Jaccard coefficient from 1, or, equivalently, by dividing the difference of the sizes of the union and the intersection of two sets by the size of the union. And we use this Jaccard distance as our second loss term:

$$L_j = 1 - J(Y,T) \quad (5)$$

where $|Y| = sum(y)$, $|T| = sum(t)$, $|Y \cap T| = sum(y * t)$, and $*$ denotes elementwise multiplication.

### 2.3. Bottleneck features with multi-scale receptive field

There is a large distribution on the size of cellular structures in the task of ET image segmentation. So the size of receptive field plays a vital part in this binary pixel-wise classification task with provided feature maps. The middle part of our network has the strongest pixel-classification ability in the whole architecture, since hidden units in center module can acquire larger receptive fields than those in lower levels due to the down-sampling process. Consequently, hidden units in the center part can obtain the most semantic information and surrounding context to determine the category of the pixel. The receptive field of our network is around 16×16 because of 4 times down-sampling, meaning the model can judge the pixel by its nearby 16×16 intensity values. This size of receptive field may be appropriate for structures with small sizes such as Microtubules, Ribosome and Centriole because they are tiny and the semantic information we get is enough to determine which category the pixel belongs to. However, when it comes to objects with bigger sizes, the semantic context acquired seems just not sufficient and false predictions may happen more possibly. One way to solve this problem is adding more down-sampling blocks in the decoder module, but this will bring more errors when localization as the decrease in sizes of bottleneck feature maps. The training burdens will be heavier because of more parameters, which may also cause overfitting as there is little training data available. Moreover, predictions for small objects or structures with large receptive field may not perform that well as too much contextual information might cause confusion to the classifier.

Our method is applying dilated convolution filters with different scales(dilation_scale = 0, 2, 4, 8) to the encoded features in parallel, thus increasing the receptive fields of hidden units in center module by 1, 2, 4, 8 times separately. We didn't use large atrous rates like ASPP in DeepLab because lots of pixels are to be padded, importing too much noise to the feature maps and causing additional loss to the final prediction, which is what we are not expecting to see in biomedical particularly ET image segmentation. Therefore we would rather lose some larger global semantic information to reduce the loss of precision. The global average pooling is also dropped, as an unsampling operation with a stride of 16 after the global average pooling layer will definitely cause some deviation to the encoded features. Then we stack them together with the input encoded feature maps and convolve the feature stack containing feature units with various kinds of receptive field sizes, making the bottleneck features learn well on both larger and smaller cellular structures.

### 2.4. Refinement of pixel spatial localization with multi-level context fusing

Due to the crowded and complex cellular environment including diverse microcellular structures and tiny tissues in ET images, precise segmentation is extremely needed so that the pathologists and biologist can analyze and make a diagnosis based on it. Therefore, the challenge is how to get precise localization results. Since most CNN-based segmentation networks use only the output of the last layer as score map for prediction, which is relatively coarse for precise segmentation details. To solve this problem, SegET uses decoded output features of every decoder block and stacks them together as multi-level context. We apply this contextual information as a combination of multiple levels of semantic abstraction and a makeup to the output of the last decoder layer as there will be a recovery loss every time an interpolation is applied. By fusing the contextual information in the higher-level decoder blocks, the deviation of localization is reduced to some degree. Therefore, there are two steps of contextual feature fusing, the fusing between higher-level decoder features and the fusing between integrated higher-level decoder features and the output feature map of the last decoder block. Meanwhile, 2 convolution layers would follow the concatenated feature stacks to refine the integrated results. One other way to do this is concatenating the outputs of every higher decoder block directly to the last feature map. We didn't do this as it will reduce the weight of feature maps of the last layer which we mainly based on for prediction and it will cost more memory space as every feature map has to interpolated to the original input size. Improvement in more fine-grained localization is observed in our experiments.

## 3. Experiments and results

### 3.1. Dataset and preprocessing

We adopt the Electron Tomography of the CTL Immunological Synapse(Project: P1694) dataset from Cell Image library [18], a searchable database, an archive of cellular images and a repository for microscopy data. The dataset contains five categories of cellular structures including synaptic cleft(Synapse), microtubules(MTs), centriole, lytic granules(Granules) and Golgi apparatus(Golgi). The microscopy is pictured by slow scan cooled 2K CCD camera from blood tissue sections and the masks are manually annotated in the pixel-level by human biologists.

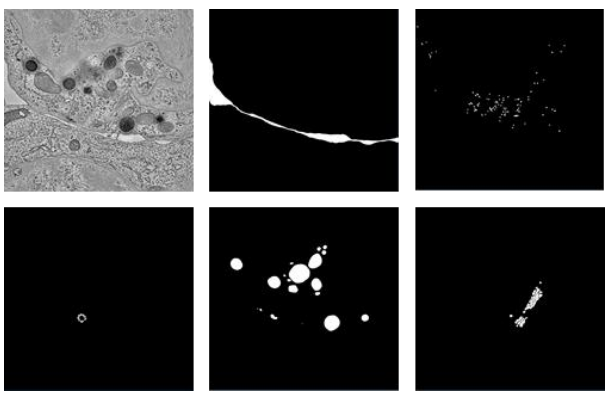

Figure 3. The original image(top left), mask of Synapse (top middle), mask of MTs(top right), mask of centriole(bottom left), mask of Granules(bottom middle), mask of Golgi(bottom right) of the $39^{th}$ slice from our dataset MRC file.

The tomographic reconstruction data is stored as an MRC file (int, 8 bit), so as each segmentation mask of those five cellular structures. Therefore, we get 6 MRC files as our whole dataset for training and evaluating. Every MRC file no matter the reconstruction data or segmentation mask is a ET image stack made up of 76 slices of 2D image with thickness of 0.15 μm, each measuring 2048px×2048px with pixel size of 2.255 nm/pixels. We hold out every 1 out of 5 images from the data and mask stack as our validation set, which takes around 20% of our total dataset. The software tool Eman2 [19] is applied when transforming MRC files into numpy arrays which can be processed directly by Python. However, each single slice of data is too large to be fed into our neural network, and the amount of training data is far too little to finish our prediction task. To solve this problem, each data slice was segmented into 512×512 patches using a sliding window with a stride of 256px, thus obtaining 7×7×60 data and mask patches for each category of cellular object to train, and 7×7×16 data and mask patches for each class of cellular structure to evaluate.

### 3.2. Implementation details

The proposed method was implemented with Python2.7 under the opensource framework of Keras architecture based on Tensorflow library. We trained the model with our training set and tested our method with the validation set on 5 distinct categories of Electron Tomography volumes separately. The hyperparameters and processing methods differ between various categories of objects. The training time and inferring time on our dataset also took distinct time spent with 24 2.10GHz Intel(R) Xeon(R) CPU E5-2620 v2 CPU and a NVIDIA Tesla K40c GPU. We use proposed BCE + Jaccard loss as our custom loss function for all training process. The metrics for evaluating the training and testing performance is Mean Intersection over Union(mIOU), which will be discussed in next part in details. The final segmentation result with multi-class were integrated by inference score maps of 5 distinct models trained for own category.

**Synapse segmentation:** For synaptic cleft segmentation, we directly fed our Synapse training and validation sets which have been patched already into proposed network. Training process was implemented with standard BP algorithm optimized by Adam [20] optimizer(initial learning rate = 1e-4, learning rate decay = 1e-6). We run for about 38 epochs with a mini-batch size of 12, monitored by an early stopper(patience=8), a model checkpointer which only saves the best model till now in the given metric, and a learning rate reducer (factor=0.5, patience=3). These callback functions are all monitored by the mean IOU value of the validation set in each epoch. And it takes 450s~550s to update the whole parameters for one epoch.

**MTs segmentation:** The main challenge for microtubules(MTs) is pixel imbalance between positive and negative pixels since the size of foreground structure is tiny and most pixels among the data patch serve as background pixels, which is obvious. The average ratio of background pixels / foreground pixels, which we call average B/F ratio, in all training data patches can come up to 508.7. The prediction results for testing data will be all background at the end of the training process even if we're using the Jaccard distance loss in our total loss, as the amount of negative pixels is way too much than the positive ones that the network will hardly pay any attention to the accuracy of the foreground pixels classification since they make little contribution to the total loss. Therefore, we use sample weights to improve the importance of positive pixels. One way to solve the problem is applying class weight, multiplying all the foreground pixel values by average BF ratio. But this method is not suitable for most tasks in ET image segmentation as the distribution of one kind of cellular structure could be very aggregated. The centralized distribution of foreground pixels causes a problem that the positive pixels are mainly distributed in a few data patches, while others contain only a small amount of positive pixels like tens or hundreds of them. It is obvious that the positive pixels in data patches with more concentrated foreground objects are overweight and those in data patches with few foreground pixels are underweight heavily. Therefore, we use an adaptive sample-wise weighting method by introducing weight matrix for every training sample, and the B/F ratio of that very training data patch will be the weight of the foreground pixels of that single training sample.

The training process of the model for MTs segmentation was also implemented with standard BP algorithm optimized by Adam optimizer(initial learning rate = 1e-3, learning rate decay = 1e-5). The upper limit for B/F ratios of weight matrices is set to 2000 in case that there are only very few foreground pixels in some training patches and the weights would be extremely huge, thus possibly causing overfitting and other deviation problems. We run for about 300 epochs with a mini-batch size of 12, monitored by an early stopper(patience=8), a model checkpointer which only saves the best model till now in the given metric, and a learning rate reducer(factor=0.5, patience=3). It takes 450s~500s training time spent for one epoch.

**Centriole segmentation:** The pixel unbalanced situation is even more serious as there is only one Centrioles in a single slice of data, which means the foreground pixels are very centralized and a few training data patches contain most of the foreground pixels. This is also not a good case for training as the overall influence and weight of foreground pixels are in a low level besides of the weight of foreground pixels in a single patch. So some tricks are used for the Centriole training set. Besides using our adaptive sample-wise weighting method to weight the foreground pixels, we copy the positive patches which contains foreground pixels by 4 times and rejoin them into the training set, getting more proportions of data that contains foreground pixels.

We optimize the training process of the model for Centriole segmentation by Adam optimizer(initial learning rate = 2e-3, learning rate decay = 1e-6). The upper limit for B/F ratios of weight matrices is set to 2000. We run for about 300 epochs with a mini-batch size of 12. An early stopper(patience=10), a model checkpointer which only saves the best model till now in the given metric, and a learning rate reducer(factor=0.5, patience=3) are applied to monitor the training process. About 630s training time will be spent for one epoch.

**Granules segmentation**: We optimize the training process of the model for Granules segmentation by Adam optimizer(initial learning rate = 2e-3, learning rate decay = 1e-5). We run for about 60 epochs with a mini-batch size of 12. An early stopper(patience=5), a model checkpointer which only saves the best model till now in the given metric, and a learning rate reducer(factor=0.5, patience=2) are applied to monitor the training process. Since there is enough ratio of positive pixels in the training set as the Synapse, no weight matrix is applied to the samples. 420s~450s training time would be expected for one epoch.

**Golgi segmentation**: The positive training patches of Golgi structure are copied by 2 times, after which the patches containing foreground pixels can take up to around 40%. Still, training loss of the model is optimizer by Adam(initial learning rate = 2e-3, learning rate decay = 1e-5). We run for about 124 epochs with a mini-batch size of 12. An early stopper(patience=5), a model checkpointer which only saves the best model till now in the given metric, and a learning rate reducer(factor=0.5, patience=2) are applied to monitor the training process. And the upper limit for B/F ratios of weight matrices is set to 1000.

### 3.3. Evaluation and comparison

The evaluation criteria we use is Mean Intersection over Union(mIOU). Supposing there are k+1 categories of pixel classes(from $C_0$ to $C_k$), the $p_{ab}$ denotes the number of pixels classified to class b which should belong to class a. One pixel is reckoned as foreground if the output value after activated by sigmoid function is greater than 0.5. Usually, the Pixel Accuracy would be used for classification tasks. This metric reflects the ratio of pixels that have been classified correctly. However, as there are mainly background pixels in all datasets, the pxAcc will be of less importance for evaluation of this task since the background pixels are more easily to be classified correctly. Finally pxAcc of all models would achieve a relatively high level. Thence, we need another strong and effective evaluation metric, Mean Intersection over Union, which defines as:

$$\text{mIOU} = \frac{1}{k+1}\sum_{i=0}^{k}\frac{p_{ii}}{\sum_{j=0}^{k}p_{ij}+\sum_{j=0}^{k}p_{ji}-p_{ii}} \quad (6)$$

where $p_{ii}$, $p_{ij}$, $p_{ji}$ still denotes the number of true positives, false positives, and false negatives, respectively. Basically, mIOU is for measuring similarity between inferring results and GT with respect to foreground and background sets.

We compared our architecture with several baseline methods mentioned before:

**FCN**: we use best performed FCN-8s as our FCN baseline. Complicated decoder is not applied as the training data is not enough for decoders like resNet152 or InceptionV3 which may cause seriously overfitting. Therefore, the decoder of our FCN-8s network is transformed into VGG16 type and it suits well. For the upsampling part, we basically maintain what it was except some parameters of conv filters are changed to fit the dataset and memory space.

**SegNet**: We retain the basic structure of original Segnet with 5 blocks for downsampling and symmetrical 5 blocks for upsampling. The amount of convolution filters of each layer is modified to fit the dataset and memory space.

| Method | Synapse | MTs | Centriole | Granules | Golgi | average |
|---|---|---|---|---|---|---|
| FCN-8s | 0.8469 | 0.6690 | 0.8287 | 0.8011 | 0.7081 | 0.7708 |
| SegNet | 0.8164 | 0.5817 | 0.8060 | 0.8576 | 0.7535 | 0.7630 |
| U-Net | 0.6765 | 0.6616 | 0.8366 | 0.9096 | 0.8020 | 0.7773 |
| DeepLabv3+ | 0.9132 | 0.6933 | 0.7280 | 0.9163 | 0.7171 | 0.7936 |
| *SegET* | 0.9322 | 0.7676 | 0.8541 | 0.9475 | 0.8687 | 0.8740 |

Table 1. The segmentation mIOU scores of different methods on separate and total cellular structures in our ET dataset

**UNet**: The basic structure of original UNet is maintained, and of course, the number of convolution filters of each layer is changed. Batch Normalization and ReLU layers are added every time a non-linear operation is applied, improving the training accuracy and efficiency.

**DeepLab v3+**: Like our FCN baseline model, the original encoder of DeepLab v3+ like resNet101 or Xception are not used to prevent the network from overfitting. Meanwhile, we adapt a VGG19-like encoder which contains 23 conv layers with 8 blocks, outperforming all other kinds of encoders we have tried. The number of convolution filters for each layer is also modified for better performance.

The hyperparameters for training of our baseline methods are the same as the proposed method by default, unless there are other hyperparameters settings which can achieve better performance on that baseline model. The training process will not stop until the given metric doesn't improve or overfitting happens. The final performance by our evaluation metric proved that the proposed method outperforms them on all cellular structures, which is demonstrated in table 1. For illustrating the efficacy of the proposed method, some segmentation results of testing data are shown in Figure 4. We can observe that SegNet and U-Net performed not so well on Synapse, due to their small convolution kernels which couldn't provide larger receptive field on such long-shape structure. It can also be observed that FCN-8s and DeepLabv3+ have poor performance on structures like Golgi with some complex borders since their decoder are relatively simple for details recovery and refinement. In the end, it is showed that our method has a strong effect on segmentation of all cellular

## 4. Conclusion

In this paper, we propose a single end-to-end deep fully-convolutional semantic segmentation architecture with a novel encoder, center and decoder part which fully exploits contextual information and reduces the loss of details from objects in ET images. Meanwhile, we also come up with a loss function for training, and some methods for the network to suit this kind of ET dataset better in the training process like adaptive sample-wise weighting. We applied our method to our dataset Electron Tomography of the CTL Immunological Synapse from Cell Image library, and the evaluation results demonstrate that the proposed method outperforms all baseline semantic segmentation methods with their best settings in all cellular structures among the dataset. In the future work, we may use more ET datasets and try data augmentation to further enlarger the capacity of the network. Threshold optimization might be applied. Some other advanced post-processing techniques would be used for further refinement of segmentation results.

## 5. Acknowledgements

This research is supported by the Strategic Priority Research Program of the Chinese Academy of Sciences Grant No. XDA19020400, the National Key Research and Development Program of China (No. 2017YFE0103900 and 2017YFA0504702), the NSFC projects Grant (No. U1611263, U1611261, 61472397, 61502455 and 61672493) and Special Program for Applied Research on Super Computation of the NSFC-Guangdong Joint Fund (the second phase).

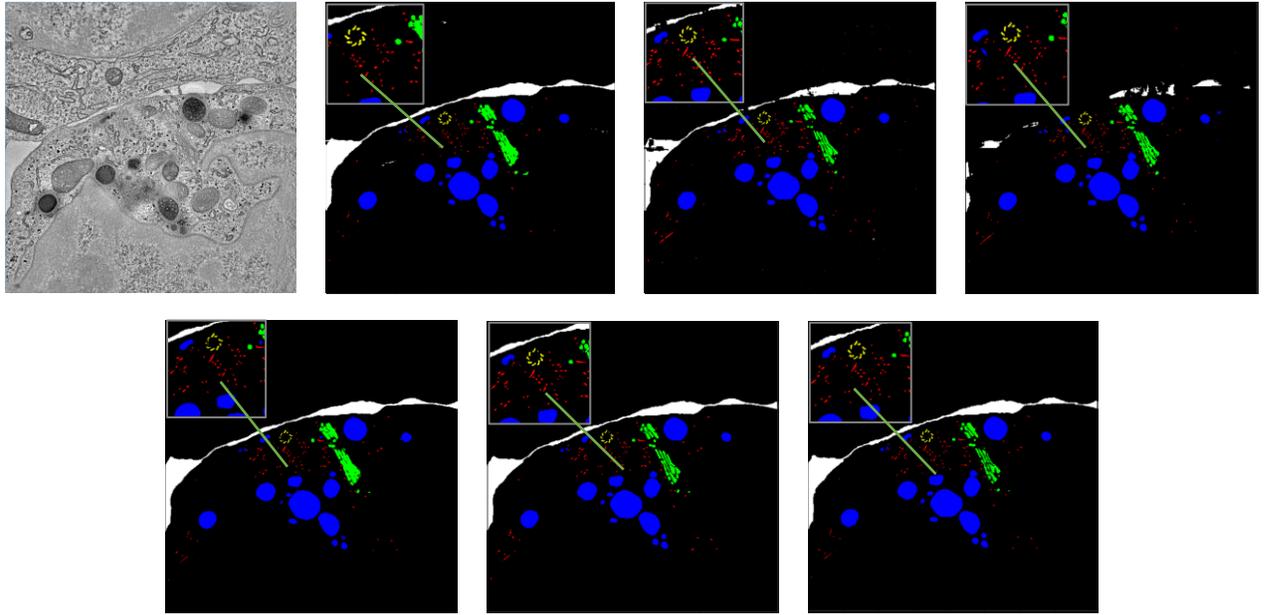

(a) Original image and segmentation results of slice 7(the 8th slice) in test set.

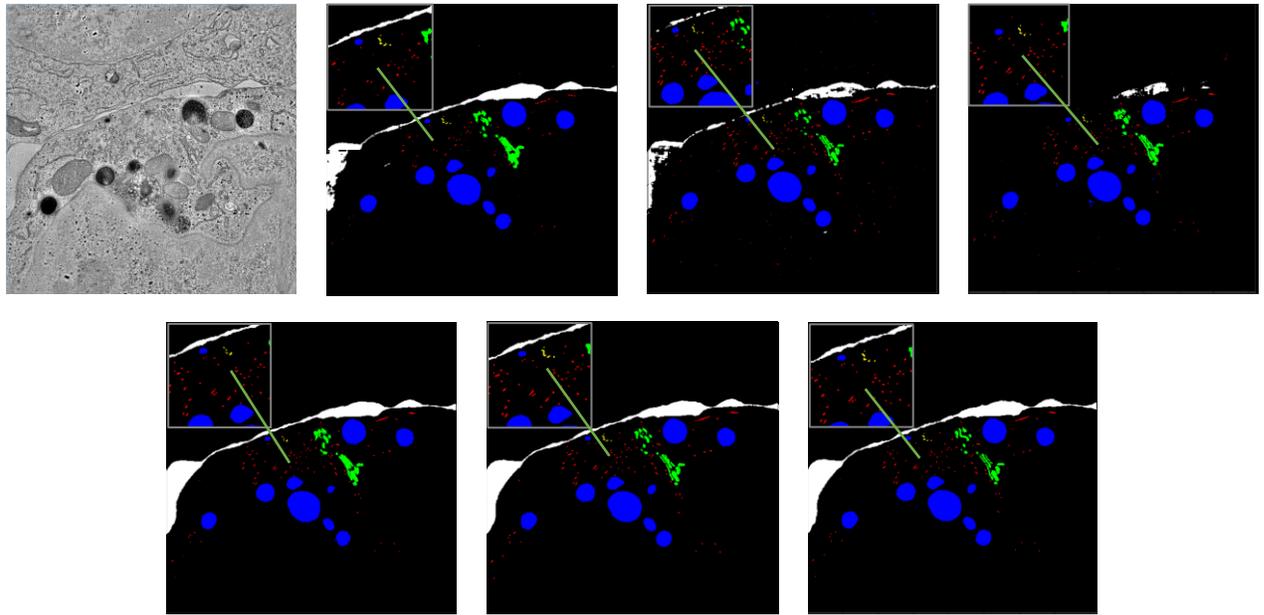

(b) Original image and the segmentation results of slice 13(the 14th slice) in test set.

Figure 4. Segmentation results of **(a)** the 8th slice in test set and **(b)** the 14th slice in test set separately. Images from top left to bottom right in turn: original images, segmentation results by FCN-8s, segmentation results by SegNet, segmentation results by U-Net, segmentation results by DeepLabv3+, segmentation results by proposed method, and ground-truth masks. Different colors denote individual cellular objects(white for Synapse, red for MTs, yellow for Centriole, blue for Granules, green for Golgi and black for background). Inside the gray square box at the top-left corner of each image is the designated area that has been zoomed in to show the details clearly.